\documentclass[runningheads]{llncs}

\usepackage{eccv}

\usepackage{eccvabbrv}

\usepackage{graphicx}
\usepackage{booktabs}

\usepackage{subcaption}
\usepackage{lipsum} 
\usepackage{multirow}
\usepackage[marginal]{footmisc}

\usepackage[accsupp]{axessibility}  

\usepackage[pagebackref,breaklinks,colorlinks,citecolor=eccvblue]{hyperref}

\usepackage{orcidlink}
\usepackage{fontawesome}

\begin{document}


\title{Cross-Modality Gait Recognition: Bridging LiDAR and Camera Modalities for Human Identification}

\titlerunning{Cross-Modality Gait Recognition}



\author{Rui Wang\inst{1}$^\star$\and
Chuanfu Shen\inst{1,2}\thanks{Equal contributions.}\and
Manuel J. Marin-Jimenez \inst{3}
\and
\\George Q. Huang\inst{4}
\and
Shiqi Yu\inst{1}\textsuperscript{(\faEnvelopeO)}
}

\authorrunning{R.Wang et al.}

\institute{Southern University of Science and Technology \and
The University of Hong Kong \and University of Cordoba \and The Hong Kong Polytechnic University}
\maketitle

\begin{abstract}
Current gait recognition research mainly focuses on identifying pedestrians captured by the same type of sensor, neglecting the fact that individuals may be captured by different sensors in order to adapt to various environments. A more practical approach should involve cross-modality matching across different sensors. Hence, this paper focuses on investigating the problem of cross-modality gait recognition, with the objective of accurately identifying pedestrians across diverse vision sensors. We present CrossGait inspired by the feature alignment strategy, capable of cross retrieving diverse data modalities. Specifically, we investigate the cross-modality recognition task by initially extracting features within each modality and subsequently aligning these features across modalities. To further enhance the cross-modality performance, we propose a Prototypical Modality-shared Attention Module that learns modality-shared features from two modality-specific features. Additionally, we design a Cross-modality Feature Adapter that transforms the learned modality-specific features into a unified feature space. Extensive experiments conducted on the SUSTech1K dataset demonstrate the effectiveness of CrossGait: (1) it exhibits promising cross-modality ability in retrieving pedestrians across various modalities from different sensors in diverse scenes, and (2) CrossGait not only learns modality-shared features for cross-modality gait recognition but also maintains modality-specific features for single-modality recognition.



\end{abstract}


\section{Introduction}
\label{sec:intro}

\begin{figure}[t]

\centering
\includegraphics[width=0.9\linewidth]{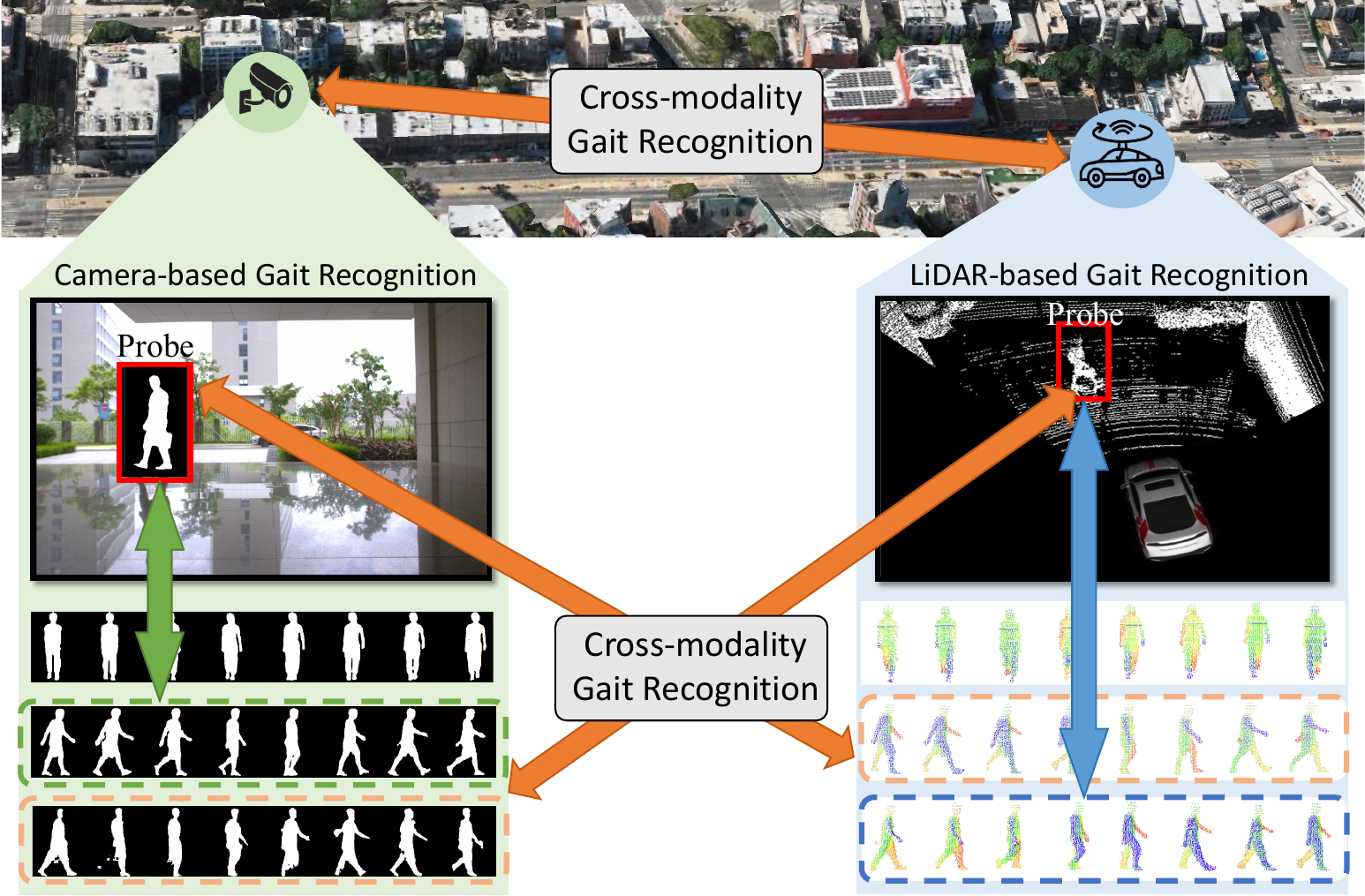}

\caption{
Comparison between single-modality and cross-modality gait recognition. The majority of methods concentrate on pedestrian retrieval within single-modality settings, where both probe and gallery subjects are captured by the same visual sensor (\textcolor{green}{green} arrow for camera-based, \textcolor{blue}{blue} arrow for LiDAR-based gait recognition). Our objective is to expand recognition across different sensors, such as LiDAR and RGB camera, as illustrated by the \textcolor{orange}{orange} arrow.
}
\vspace{-3em}
\label{fig:probegallery}

\end{figure}

Gait recognition~\cite{nixon2006automatic,chao2019gaitset, liao2020model,liang2022gaitedge,shen2023lidargait,han2022licamgait} has gained significant attention in recent years, because of its potential for long-distance pedestrian retrieval based on walking patterns. Thanks to the advancements in deep learning, gait recognition has achieved remarkably high performance across various challenging scenarios. For instance, state-of-the-art methods~\cite{fan2023exploring} have gained a 90\% Rank-1 recognition accuracy in well-known in-the-lab datasets and approximately 80\% Rank-1 accuracy in in-the-wild datasets. Current research operates primarily on camera-based gait recognition and formulates gait recognition as a single-modality matching problem,~\ie probe and gallery samples during testing maintain a consistent format in modality.

However, single-modality camera-based gait recognition may face limitations in real-world scenarios with poor or unavailable lighting conditions. To address this limitation, many recent work~\cite{10448817,shen2023lidargait,han2022licamgait,benedek2016lidar} explore the utilization of LiDAR for robust perception regardless of bad illuminations. While both LiDAR-based and camera-based gait recognition may perform satisfactorily in single-modality matching settings, a more realistic setting should involve cross-modality matching across different sensors with heterogeneous modalities. As illustrated in Fig.~\ref{fig:probegallery}, the person-of-interest can be commonly appeared in different scenarios and captured by different visual sensors, such as visual cameras and LiDAR sensors. Since people live in the 3D wild world, no single visual sensor can handle all different environments. This means we need different sensors that can capture different kinds of information to adapt to various real-world situations. Thus, it is essential to study cross-modality gait recognition for real-world applications. 

To the best of our knowledge, there are no works on gait recognition between RGB cameras and LiDAR sensors in the literature. As illustrated in Fig.~\ref{fig:probegallery}, gait representations obtained from RGB cameras and LiDAR sensors are inherently distinct and possess different formats. Camera-based gait representations typically take the form of silhouettes, organized as two-dimensional grids, whereas LiDAR sensors generate three-dimensional point clouds characterized by sparsity. The challenges of LiDAR-Camera cross-modality gait recognition lie in two aspects. (1) It is challenging to learn a modality-shared metric space from two inherently distinct gait representations,~\ie 3D point clouds and silhouettes, containing varying levels of semantic information. (2) At the same time, it is also significant to preserve identity-discriminative metric space rather than only bridge the gap between modality features.

In this paper, we propose the first cross-modality gait recognition framework, named CrossGait. Specifically, Our method employs a straightforward feature alignment technique to bridge the gap between LiDAR point clouds and camera silhouettes. Given the dramatic difference between these modalities, directly aligning point clouds and silhouettes in a shared feature space remains challenging. Thus, we introduce the Prototypical Modality-Shared Attention Module (PMAM) to emphasize the extraction of common features among the learned features of different modalities. 
Additionally, we propose a cross-modality feature adapter to synchronize features from disparate modalities into a cohesive, modality-shared feature space, thereby bridging the considerable gap between the distinct data modalities.

To summarize, the main contributions of this work are as follows:
(1) To the best of our knowledge, this study represents the first exploration of cross-modality gait recognition involving both the LiDAR and visual camera. 
(2) We proposed a simple yet effective framework, named CrossGait, for cross-modality gait recognition. CrossGait demonstrates its ability to retrieve pedestrians across various modalities from different sensors in diverse scenes. Furthermore, it consistently delivers satisfactory performance even in single-modality settings.
(3) The CrossGait method performs well when it deals with data captured from different sensors, showing superior results in cross-modality scenarios. Moreover, it can also handle data from the same device but in various formats like human parsing, silhouettes, and skeleton maps.

\section{Related Work}
\subsection{Gait recognition}
Gait recognition can be classified based on the type of sensors into 2D gait recognition~\cite{chao2019gaitset,liao2020model,liao2017pose,teepe2021gaitgraph,liang2022gaitedge,zheng2022gait} and 3D gait recognition~\cite{shen2023lidargait,benedek2016lidar,han2022licamgait}.
In 2D gait recognition, the primary source of gait representations is RGB cameras. These cameras capture images with color and texture information in a 2D grid format. However, this data can obscure gait analysis since it includes irrelevant features. To address this, researchers commonly convert gait sequences from RGB images into silhouettes, parsing, or skeleton~\cite{chao2019gaitset,fan2020gaitpart,huang2021context,lin2022gaitgl,lin2021gait, fan2022learning, fan2023exploring,liao2020model,liao2017pose,teepe2021gaitgraph,an2018improving,an2020performance,wagg2004automated}. Among these 2D gait representations, silhouettes have emerged as the dominant choice over the past two decades due to its simplicity and effectiveness.
However, there is a fundamental drawback to utilizing camera-based representations for 2D gait recognition. These representations are sensitive to lighting conditions and visual ambiguity, which limits their applicability in real-world environments.
To this end, many recent work~\cite{shen2023lidargait,benedek2016lidar,han2022licamgait} employ LiDAR sensors to provide robust pedestrian perception in complex environments. Unlike cameras, LiDAR can precisely capture 3D geometry in both large-scale indoor and outdoor settings without being affected by lighting conditions. By using 3D geometry for gait recognition, LiDAR has expanded gait recognition into the realm of 3D environments, achieving superior results compared to traditional camera-based 2D gait recognition, and showing promising performance across various challenging scenarios~\cite{shen2022gait}.

In previous research, subjects were typically studied under ideal conditions, where probe and gallery sets were obtained from the same sensor settings. However, real-world applications often require the use of different sensors in various scenarios due to economic and practical considerations. For example, visual cameras are commonly utilized in residential areas due to their cost-effectiveness, while LiDAR sensors are widely used in robotics and autonomous vehicles. As illustrated in Fig.~\ref{fig:probegallery}(a), individuals of interest may be observed in both residential regions and on the streets. Therefore, in this work, we investigate the cross-modality task in gait recognition to accommodate pedestrians appearing under different sensors and scenarios, thereby expanding the range of applications.

\subsection{Cross-Modality Modelling}

\begin{figure*}[t]
  \centering
   \includegraphics[width=0.9\linewidth]{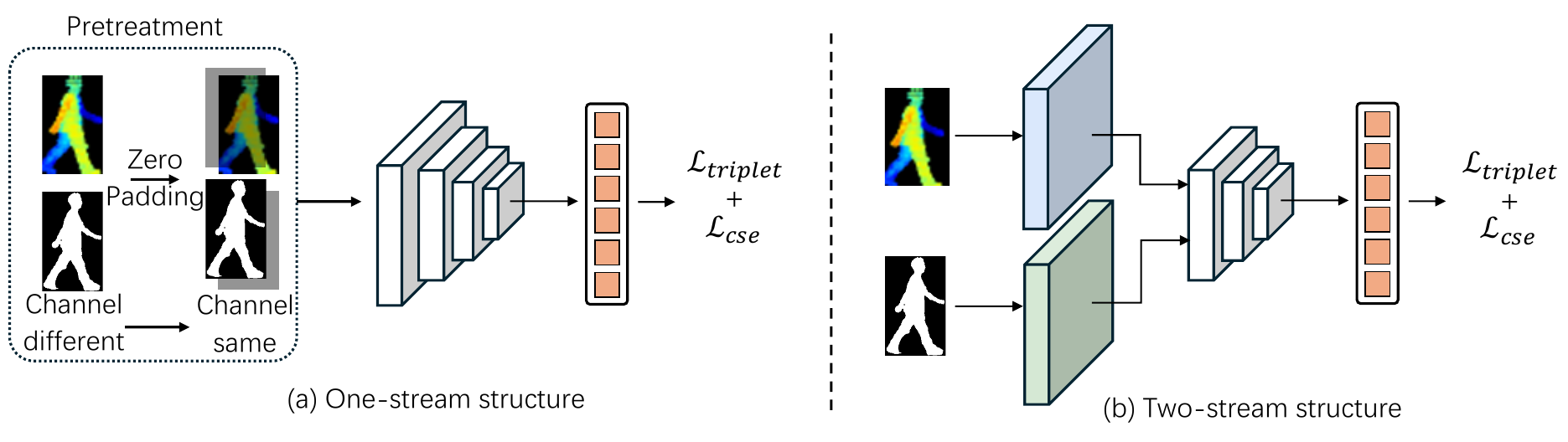}
\vspace{-1em}
    \caption{One-stream and two-stream cross-modality framework.}
    \label{fig:twokind}
\vspace{-2.5em}
\end{figure*}

\noindent\textbf{Cross-modality Person Re-identification.} 
Although there is limited research on cross-modality retrieval in gait recognition, a related field known as person re-identification (ReID) has extensively investigated cross-modality problems, particularly focusing on visible-infrared person re-identification (VI-ReID)~\cite{wu2017rgb,xiang2019cross,nguyen2017person,ye2018hierarchical,wang2019learning}.
Wu \textit{et al.}~\cite{wu2017rgb} introduced the pioneering VI-ReID framework called Zero Padding, which adds a zero channel to each grid-based input for representation alignment, as illustrated in Fig.~\ref{fig:twokind}. Conceptually, Zero Padding can be viewed as a technique that employs input-level feature alignment. However, relying on manually crafted engineering to forcibly align different modalities into a common space at the input stage often results in poor performance.
Later on, more VI-ReID frameworks have increasingly employed the two-stream architecture~\cite{yu2023modality,fang2023visible,wu2023learning,wu2021discover,lin2022learning} with better results.  As illustrated in Fig.~\ref{fig:twokind}(b), the two-stream architecture separately encodes input data from domains and learns a shared cross-modality encoder. 
Extensive experiments have demonstrated the superiority of the two-stream architecture, which incorporates intermediate-level feature alignment.
While VI-ReID has made significant strides in cross-modality retrieval, directly applying existing frameworks for cross-modality gait recognition is not promising. The main difference lies in the type of data they handle. In cross-modality person re-identification, the focus is on matching visible and infrared images, both of which are in 2D grid format. On the other hand, cross-modality gait recognition deals with sparse point clouds and silhouettes, presenting a more challenging task. To compare with existing methods, we adopt the classical two-stream architecture as the baseline for our approach.

\noindent\textbf{Multi-modality Representation Learning.} 
Multi-modality representation learning can be broadly divided into two main strategies. 
The first strategy focuses on complementing information across different modalities to enhance performance in various tasks.
This approach is prevalent in many multi-modality studies, such as those combining image and text modalities~\cite{chen2020uniter,li2021align,li2019visualbert,li2020oscar,lu2019vilbert,tan2019lxmert,radford2021learning}, where the goal is to leverage the strengths of one modality to compensate for the limitations of another. Notably, in the context of gait recognition~\cite{han2022licamgait,castro2020multimodal,zhao2021multimodal}, utilizing multi-modal methods has been shown to significantly improve performance by integrating features across different modality data.
The second strategy aims to construct a unified feature space that can effectively represent and integrate information from diverse modalities. 
Such as CLIP~\cite{radford2021learning}, utilizes image and text encoders to produce a unified image/text representation for each image-text pair enhancing its zero-shot generalization capacity. 
The release of CLIP catalyzed a surge in research exploring the interplay between images and text~\cite{patashnik2021styleclip,gao2022open,gu2021open,li2022grounded,chen2021multimodal,yan2022let}, promoting advancements in areas such as text-driven image retrieval, open vocabulary object detection, and visual-linguistic grounding. These endeavours share the goal of crafting a comprehensive representation framework to bridge multiple modalities.

In our cross-modality gait recognition study, we draw inspiration from CLIP and develop a unified identity representation space capable of accommodating different modalities. We extract features from gait sequences of two distinct modalities using independent feature extractors without parameter sharing, which enables the learning of modality-specific features for single-modality gait retrieval. Additionally, by designing a cross-modality feature adapter to encode a common shared feature space, we facilitate the learning of modality-shared features for cross-modality gait recognition.


\begin{figure*}[t]
  \centering
   \includegraphics[width=0.9\linewidth]{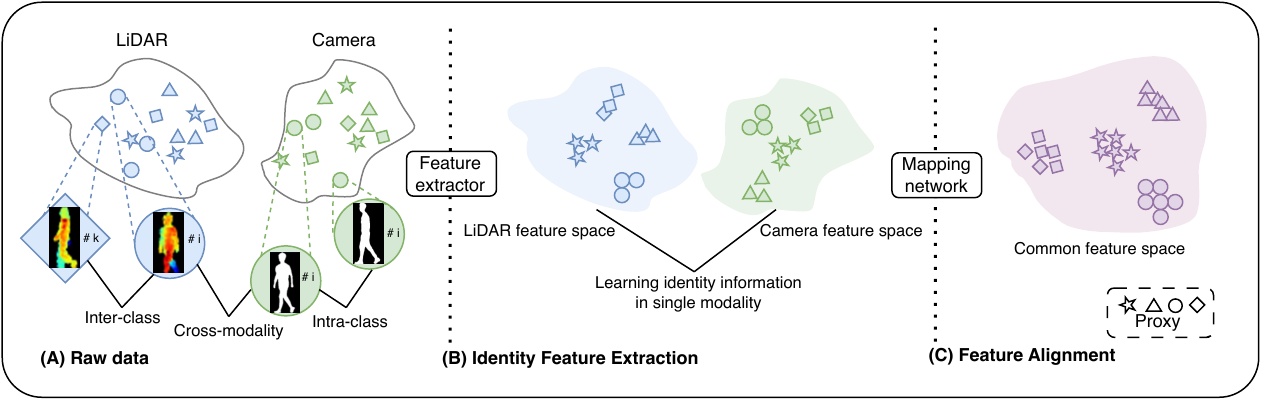}
    \vspace{-1em}
    \caption{Feature alignment strategy for cross-modality gait recognition. (A) illustrates the complexities of cross-modality gait recognition, including modality and intra-modality variations in raw data (represented by two identities $i$ and $k$). (B) represents the process of feature extraction into two isolated modality-specific feature spaces. (C) aligns two isolated spaces into a unified modality-shared space.}

   \label{fig:method}
   \vspace{-2em}
\end{figure*}

\section{CrossGait}
There are significant challenges in cross-modality gait recognition: (1) learning a modality-shared feature space to enable retrieval between two distinct modalities, and (2) simultaneously maintaining a modality-specific feature space to enable retrieval within the same modality. To address these challenges, we employ a two-stage gait recognition framework, named CrossGait, which is capable of both single-modality and cross-modality retrieval. As illustrated in Fig.~\ref{fig:method}, in the first stage, modality-specific feature spaces are formulated by discriminative metric learning separately for each modality, following traditional single-modality gait recognition methods. To bridge the gap between these separate modality-independent feature spaces, we propose a prototypical modality-shared attention module and a cross-modality feature adapter to connect the modality-specific spaces and formulate a modality-shared feature space for cross-modality retrieval.




\subsection{Pipeline}
\begin{figure*}[t]
  \centering
   \includegraphics[width=0.85\linewidth]{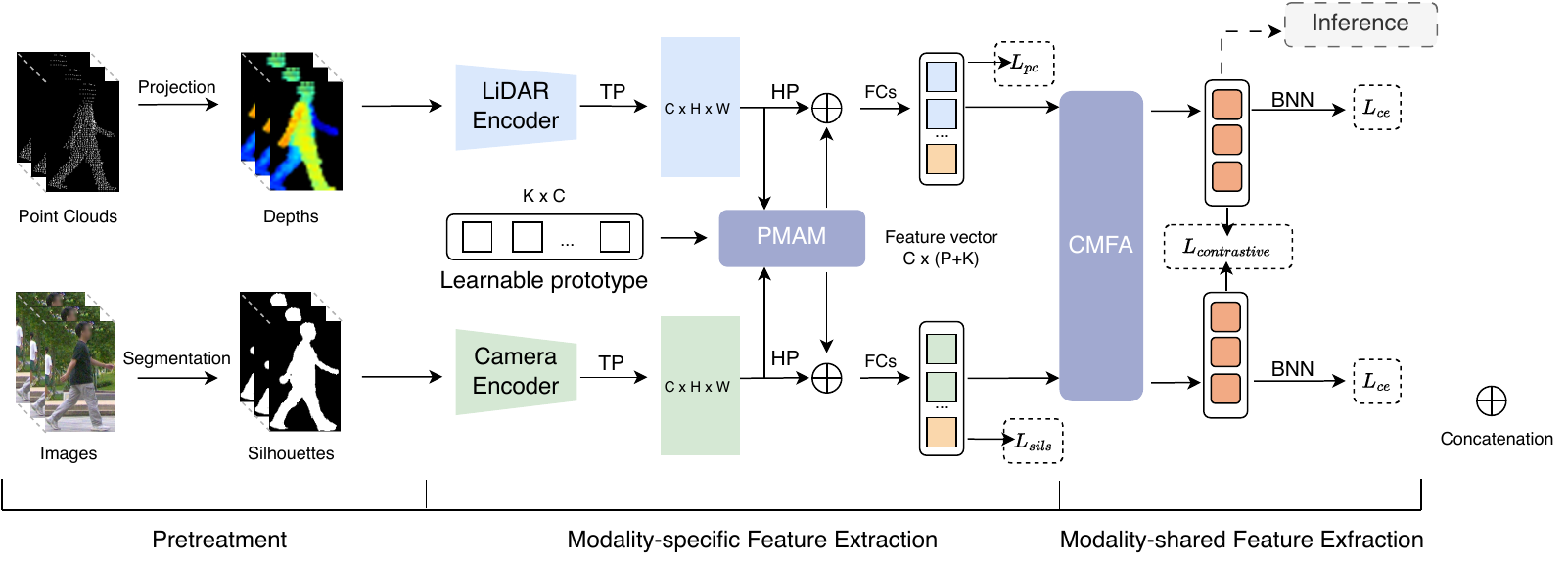}
   
 
   \vspace{-2ex}
   \caption{The pipeline of our proposed CrossGait. Our method involves learning both modality-specific features and modality-shared prototypes using a Prototypical Modality-Shared Attention Module (PMAM), followed by a Cross-modality Feature Adapter to transform the modality-specific features into a unified feature space that encodes the modality-shared features.}
   \label{fig:pipeline}
   \vspace{-4ex}
\end{figure*}

This paper primarily focuses on cross-modality gait recognition between LiDAR point clouds and camera images. Our proposed CrossGait is capable of processing gait sequences in terms of either LiDAR point clouds or camera images, as illustrated in Fig.~\ref{fig:pipeline}. Given input LiDAR point clouds $P = \{P^i | P^i\in\mathbb{R}^{N\times 3} \}^t_{i=1}$ or RGB images $I = \{I^i | I^i \in \mathbb{R}^{3 \times H \times W}\}^T_{i=1}$, each modality of gait representations needs to be preprocessed into corresponding representations. LiDAR point clouds are projected into depth maps $D = \{D^i | D^i \in \mathbb{R}^{C \times H \times W}\}^t_{i=1}$ because there is no effective point cloud encoder for gait recognition utilizing raw point clouds, while the RGB images are segmented and aligned following silhouette-based gait recognition methods $S = \{S^i | S^i \in \mathbb{R}^{H \times W}\}^T_{i=1}$, where $T$ represents the number of frames in the sequence of the camera-based modality, $t$ represents the number of frames in the sequence of the LiDAR-based modality, $N$ indicates the number of points in each frame of the point cloud, $H$, $W$, and $C$ denote the height, width, and channels of the silhouettes, respectively. Next, we obtain modality-specific features for each modality by employing different feature encoders on projected depths and silhouettes. We use a commonly used convolutional architecture~\cite{fan2023opengait} as the feature encoder, followed by a temporal pooling layer. To enable cross-modality matching, we introduce the prototypical modality-shared attention module to extract modality-shared features between these modalities. Lastly, we propose a cross-modality feature adapter to align learned features from each modality into a unified space. CrossGait isolates the learning of modality-specific features and modality-shared features, enabling promising single-modality matching and cross-modality matching.

\subsection{Prototypical Modality-shared Attention Module}

\begin{figure}[t]
\centering

    \includegraphics[width=0.65\linewidth]{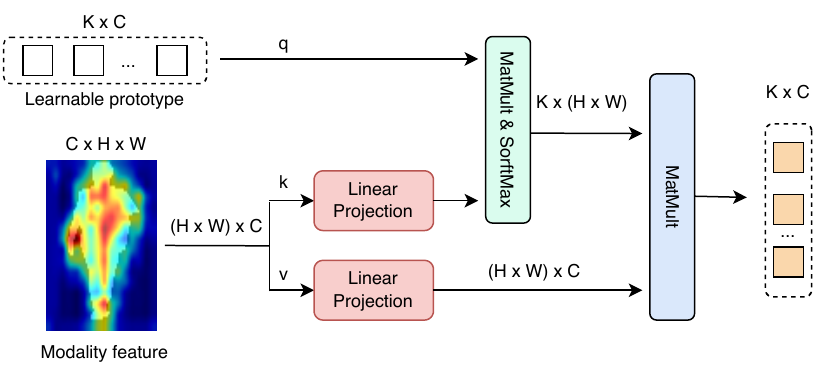}
\vspace{-1em}
    \caption{The structure of Prototypical Modality-shared Attention Module (PMAM).}
\vspace{-2em}
    \label{fig:semantic}
\end{figure}

The Prototypical Modality-shared Attention Module is introduced to extract modality-shared features between different modalities in the context of cross-modality gait recognition as illustrated in Fig.~\ref{fig:semantic}. This module is designed to capture common features among the learned features of different modalities, thereby facilitating cross-modality matching. By leveraging attention mechanisms, the module can effectively focus on relevant information from both modalities and learn representations that are useful for recognizing gait patterns regardless of the input modality. 
The Prototypical Modality-shared Attention Module initializes learnable prototypes $Q \in \mathbb{R}^{K \times C}$, which serve as references for common feature extraction across modalities, where $K$ denotes the number of prototypes and $C$ represent the channel size.
This module establishes attention relationships between the two modalities during the feature extraction stage, enabling them to focus on common features. The ($k$, $v$) pairs are generated by applying two linear projection layers on the learned modality-specific features $F$.
\begin{equation}
q = Q, \ k = FW^k,\  v = FW^v
\end{equation}
where $W^k\in\mathbb{R}^{HW \times HW}$ and $W^v\in\mathbb{R}^{HW \times HW}$ are linear projections operating at the pixel dimension, $HW$ represent the number of pixels, and $k,v\in\mathbb{R}^{HW \times C}$. The attention weights between the query $q$ and the key $k$ are derived by the inner product with a scaling operation and Softmax normalization. We then obtain refined features as the weighted sum of values $v\in\mathbb{R}^{HW \times C}$ and attention weights. This process yields the modality-shared features $F_{\text{prototype}} \in \mathbb{R}^{K \times C}$, formally represented as:
\begin{equation}
F_{\text{prototype}} = \text{Softmax}\left(\frac{qk}{\sqrt{HW}}\right)v
\end{equation}

The learned modality-shared features are then concatenated with horizontally partitioned modality-specific features, forming fused LiDAR-based features $f_D \in \mathbb{R}^{(K+p) \times C}$ and fused camera-based features $f_S \in \mathbb{R}^{(K+p) \times C}$, where $p$ indicate the part number of horizontal partitioning layer. To refine these identity features within each modality, we apply modality-specific triplet losses.
For the LiDAR-based branch, the learning process is structured as follows:
\begin{equation}
L_{\text{pc}} = \frac{1}{N_{tp+}}\sum_{\substack{a,p,n\\y_a=y_p\neq y_n}}max(m+d(a,p)-d(a,n),0)
\end{equation}
where $N_{tp+}$ denotes the number of triplets of non-zero loss terms in a mini-batch, $a, p, n$ stand for anchor, positive and negative, respectively. $d(a,p)$ and $d(a,n)$ denotes the distance between anchor-positive and anchor-negative respectively. The hyper-parameter $m$ in contrastive loss refers to the margin. For the camera-based branch, the identity information within learned features is preserved by optimizing a similar loss $L_{\text{sils}}$, but only for silhouette triplets.


\subsection{Cross-modality Feature Alignment}
In this stage, we aim to integrate features from diverse modalities into a unified identity feature space using our proposed Cross-modality Feature Adapter (CMFA) and contrastive alignment loss. Specifically, the CMFA employs part-separated fully-connected layers to transform modality-specific features into a unified embedding capable of handling cross-modality scenarios. This process can be formulated as follows:
\begin{equation}
f^{'}_{D} = W_{\text{shared}} \cdot f_D \quad \text{and} \quad f^{'}_{S} = W_{\text{shared}} \cdot f_S
\end{equation}
where $f^{'}_{D}$ and $f^{'}_{S}$ represent the transformed features in the shared feature space. $W_{\text{shared}}$ denotes the parameters of the shared fully connected layers. By applying these fully connected layers, the CMFA aligns the modality-specific features into a unified identity feature space. This approach enable the capability to recognize and compare gait patterns across different input modalities effectively. The feature alignment is achieved by optimizing contrastive alignment loss:
\begin{equation}
d = f^{'}_{D} - f^{'}_{S}
\vspace{-2em}
\end{equation}

\begin{equation}
L_{\text{contrastive}} = \frac{1}{N_{p}}\sum_{n=1}^{N_{p}}(y_nd^2+(1-y_n)max(m-d^2,0))
\end{equation}
where $N_{p}$ is the number of contrastive pairs. $y_n$ is a binary indicator that denotes whether the pair of samples $f^{'}_{D}$ and $f^{'}_{S}$ belong to the same subject ($y_i = 1$) or not ($y_i = 0$), $d$ is the Euclidean distance between the embeddings of each pair, and $m$ is a margin hyperparameter that controls the minimum distance between samples of different classes.


\subsection{Training and Inference.}
In this paper, we train CrossGait by employing a combination of multiple loss functions:
\begin{equation}
L  = L_{\text{pc}} + L_{\text{sils}} +L_{\text{ce}} + \lambda L_{\text{contrastive}}
\label{eq:opt}
\end{equation}
where $L$ represents the composite loss function, encompassing two modality-specific losses ($L_{\text{pc}}$ and $L_{\text{sils}}$), a contrastive alignment loss $L_{\text{contrastive}}$ designed to align modalities within a unified space, and a cross-entropy loss $L_{\text{ce}}$ aimed at augmenting the discriminative capability of the learned features. The hyperparameter $\lambda$ is employed to control the influence of cross-modality interactions.

During the inference, we adopt Euclidean distance as a metric to calculate the distance between probe and gallery following existing gait recognition methods~\cite{fan2023opengait,shen2023lidargait}. Our primary focus in this paper is cross-modality matching, which is evaluated through two settings. In the first evaluation protocol of cross-modality gait recognition, all LiDAR-based modalities are employed as the gallery set, while camera-based silhouettes are used as the probe. In the second evaluation protocol, the roles are reversed.



\section{Experiments}

\subsection{Datasets}
\textbf{SUSTech1K}~\cite{shen2023lidargait} is used for evaluation because it is the only publicly available dataset for gait recognition that provides both LiDAR and RGB camera modalities. The dataset consists of 25,239 sequences, capturing a diverse range of variations such as visibility, viewpoints, occlusions, clothing, carrying items, and scenes. This rich variability in the dataset ensures that models trained on SUSTech1K are exposed to a wide spectrum of real-world scenarios, enhancing their robustness and generalizability. It is divided into a training set with 250 subjects and 6,011 sequences, and a test set with 800 subjects and 19,228 sequences, offering comprehensive data for robust cross-modality gait recognition research. Given the comprehensive and diverse nature of SUSTech1K, it has been selected as the primary benchmark to study cross-modality gait recognition.

\noindent\textbf{CCPG}~\cite{li2023depth} is also utilized for evaluation, despite containing gait data captured solely by RGB cameras. This dataset allows us to extend our study to explore matching between different types of input, including silhouettes, skeletons, and human parsing. The CCPG dataset specifically focuses on challenging cloth-changing scenarios, featuring a diverse collection of coats, pants, and bags in various colors and styles. This diversity addresses the limitations of the SUSTech1K dataset, where suboptimal performance may occur under clothing variation conditions. By incorporating the CCPG dataset into our experiments, we aim to evaluate the adaptability of our method across different datasets, particularly in scenarios involving changes in clothes.

\subsection{Experimental Setting}
\noindent \textbf{Evaluation Protocol.} For cross-modality gait recognition, we draw on the evaluation criteria from visible-infrared person re-identification~\cite{nguyen2017person}, which includes \textit{LiDAR $ \rightarrow $ Camera} and \textit{Camera $ \rightarrow $ LiDAR} settings. These settings represent the direction of data flow in the cross-modality recognition process. In the \textit{LiDAR $\rightarrow$ Camera} setting, we evaluate the cross-modality matching by using all LiDAR-based data as the probe and all camera-based data as the gallery. This allows us to assess how well the system recognizes a gait using LiDAR data and matches it with corresponding camera images in the gallery.
Additionally, we explore the cross-view recognition evaluation protocol on the SUSTech1K dataset. This protocol aligns with the cross-view evaluation protocols used in CASIA-B~\cite{yu2006framework} and OUMVLP~\cite{an2020performance}. Overall, by considering these evaluation criteria and protocols, we ensure a comprehensive assessment of cross-modality gait recognition systems in diverse scenarios.

\noindent \textbf{Implementation Details.} During the training phase, we configured the batch sizes $(p,k,l)$ as (8,8,10) for LiDAR and (8,8,30) for the camera. Here, $p$ represents the number of unique individuals in the batch, $k$ denotes the number of gait sequences per individual, and $l$ represents the number of frames per gait sequence. The difference in sequence length for the two modalities is due to the varying frame rates of the LiDAR and camera sensors. We extracted input within one gait cycle at a resolution of 64 × 64 for both modalities. We used the Adam optimizer with an initial learning rate of $1\times10^{-3}$. The learning rate is reduced by a factor of 0.1 at the 30k, 40k, and 50k iterations, with a total of 60k iterations. For Triplet and contrastive alignment losses, we set the margin to 0.2 and the hyper-parameter $\lambda$ to 2. We also set the prototypical modality-shared attention module parameter $K$ to 2. To ensure fair comparisons between different methods, we followed identical training procedures for all comparable approaches. All methods were implemented using the OpenGait~\cite{fan2023opengait} codebase to ensure consistency and reproducibility.



\subsection{Comparative Results}


\begin{table*}[t]
\caption{Evaluation on SUSTech1K dataset~\cite{shen2023lidargait} under cross-modality (Bottom) and single-modality protocol (Top).}
\vspace{-1em}
\centering
\renewcommand{\arraystretch}{1.2}
\resizebox{\textwidth}{!}{%
\begin{tabular}{cccccccccccccc}
\toprule
\multicolumn{14}{c}{\textbf{Single-modality gait recognition}}                                                                                                                                                                                                                                                                                                                                                                 \\ 
\hline
\textbf{}                             & \multicolumn{1}{c|}{}            & \multicolumn{6}{c|}{Camera $\rightarrow$ Camera}                                                                                                                                      & \multicolumn{6}{c}{LiDAR $\rightarrow$ LiDAR}                                                                                                                            \\ \cline{3-14} 
Method                                & \multicolumn{1}{c|}{Publication} & \multicolumn{4}{c|}{Probe Sequence (\textit{R@1} acc)}                                                             & \multicolumn{2}{c|}{Overall}                               & \multicolumn{4}{c|}{{\color[HTML]{000000} Probe Sequence (\textit{R@1} acc)}}                                      & \multicolumn{2}{c}{Overall}                   \\
                                      & \multicolumn{1}{c|}{}            & Normal                & Clothing              & Umbrella              & \multicolumn{1}{c|}{Night}         & \textit{R@1}                & \multicolumn{1}{c|}{\textit{R@5}}        & Normal                & Clothing              & Umbrella              & \multicolumn{1}{c|}{Night}         & \textit{R@1}                & \textit{R@5}                \\ \hline
GaitBase~\cite{fan2023opengait}                              & \multicolumn{1}{c|}{CVPR2023}    & 81.4                  & 49.6                  & 75.5                  & \multicolumn{1}{c|}{25.9}          & 76.1                  & \multicolumn{1}{c|}{89.3}          & -                     & -                     & -                     & \multicolumn{1}{c|}{-}             & -                     & -                     \\
LidarGait~\cite{shen2023lidargait}                             & \multicolumn{1}{c|}{CVPR2023}    & -                     & -                     & -                     & \multicolumn{1}{c|}{-}             & -                     & \multicolumn{1}{c|}{-}             & 91.8                  & 74.6                  & 89.0                  & \multicolumn{1}{c|}{90.4}          & 86.7                  & 96.1                  \\
CrossGait                             & \multicolumn{1}{c|}{Ours}        & 75.3                  & 44.1                  & 69.9                  & \multicolumn{1}{c|}{38.3}          & 71.1                  & \multicolumn{1}{c|}{87.6}          & 90.6                  & 71.2                  & 66.5                  & \multicolumn{1}{c|}{87.0}          & 84.9                  & 95.2                  \\  \hline \hline
\multicolumn{14}{c}{\textbf{Cross-modality gait recognition}}                                                                                                                                                                                                                                                                                                                                                            \\ \hline

\textbf{}                             & \multicolumn{1}{c|}{}            & \multicolumn{6}{c|}{LiDAR $\rightarrow$ Camera}                                                                                                                                       & \multicolumn{6}{c}{Camera $\rightarrow$ LiDAR}                                                                                                                           \\ \hline
GaitBase~\cite{fan2023opengait}                              & \multicolumn{1}{c|}{CVPR2023}    & \multicolumn{1}{c}{-} & \multicolumn{1}{c}{-} & \multicolumn{1}{c}{-} & \multicolumn{1}{l|}{-}             & \multicolumn{1}{c}{-} & \multicolumn{1}{l|}{-}             & \multicolumn{1}{c}{-} & \multicolumn{1}{c}{-} & \multicolumn{1}{c}{-} & \multicolumn{1}{l|}{-}             & \multicolumn{1}{c}{-} & \multicolumn{1}{c}{-} \\
LidarGait~\cite{shen2023lidargait}                             & \multicolumn{1}{c|}{CVPR2023}    & \multicolumn{1}{c}{-} & \multicolumn{1}{c}{-} & \multicolumn{1}{c}{-} & \multicolumn{1}{l|}{-}             & \multicolumn{1}{c}{-} & \multicolumn{1}{l|}{-}             & \multicolumn{1}{c}{-} & \multicolumn{1}{c}{-} & \multicolumn{1}{c}{-} & \multicolumn{1}{l|}{-}             & \multicolumn{1}{c}{-} & \multicolumn{1}{c}{-} \\
CAJ~\cite{ye2021channel}                                   & \multicolumn{1}{c|}{ICCV2021}    & 15.3                  & 6.4                   & 13.0                  & \multicolumn{1}{c|}{2.3}           & 12.3                  & \multicolumn{1}{c|}{32.3}          & 16.4                  & 7.5                   & 7.4                   & \multicolumn{1}{c|}{2.4}           & 11.3                  & 30.1                  \\
SAAI~\cite{fang2023visible}                                  & \multicolumn{1}{c|}{ICCV2023}    & 26.5                  & 21.9                  & 23.2                  & \multicolumn{1}{c|}{3.2}           & 26.1                  & \multicolumn{1}{c|}{54.1}          & 22.4                  & 14.3                  & 14.0                  & \multicolumn{1}{c|}{5.3}           & 23.1                  & 49.5                  \\
One-stream network (Depth Silhouette)~\cite{shen2023lidargait} & \multicolumn{1}{c|}{CVPR2023}    & 23.2                  & 14.2                  & 24.7                  & \multicolumn{1}{c|}{2.4}           & 18.3                  & \multicolumn{1}{c|}{39.9}          & 18.2                  & 3.4                   & 3.4                   & \multicolumn{1}{c|}{4.7}           & 9.6                   & 22.9                  \\
One-stream network (Zero-Padding)~\cite{wu2017rgb}     & \multicolumn{1}{c|}{ICCV2017}    & 24.5                  & 14.2                  & 18.3                  & \multicolumn{1}{c|}{3.8}           & 21.0                  & \multicolumn{1}{c|}{41.9}          & 33.4                  & 9.9                   & 18.8                  & \multicolumn{1}{c|}{2.0}           & 23.1                  & 46.4                  \\
Two-stream network~\cite{wu2017rgb}                    & \multicolumn{1}{c|}{ICCV2017}    & 52.0                  & 25.1                  & 47.2                  & \multicolumn{1}{c|}{4.1}           & 43.5                  & \multicolumn{1}{c|}{69.1}          & 57.0                  & 28.7                  & 28.2                  & \multicolumn{1}{c|}{10.7}          & 45.8                  & 70.8                  \\ \hline
CrossGait                             & \multicolumn{1}{c|}{Ours}        & \textbf{62.2}         & \textbf{35.4}         & \textbf{57.8}         & \multicolumn{1}{c|}{\textbf{10.3}} & \textbf{56.4}         & \multicolumn{1}{c|}{\textbf{79.7}} & \textbf{63.2}         & \textbf{30.6}         & \textbf{38.5}         & \multicolumn{1}{c|}{\textbf{11.8}} & \textbf{53.6}         & \textbf{77.0}         \\ \hline
\end{tabular}%
\vspace{-2em}
\label{Tab:ModelComparisons}
}

\end{table*}


In Tab.~\ref{Tab:ModelComparisons}, we present a comprehensive analysis of the performance of our proposed method, CrossGait, in comparison to other cross-modality methods on the SUSTech1K dataset. To better understand the limitations and challenges of cross-modality recognition, we also report the results of single-modality gait recognition, which allows us to assess the state-of-the-art results of modalities of LiDAR point clouds and camera silhouettes.

\noindent \textbf{Single-modality Evaluation.} We compare our proposed CrossGait with two widely used baseline methods for silhouette-based~\cite{fan2023opengait} and LiDAR-based gait recognition~\cite{shen2023lidargait} in the single-modality evaluation protocol. To ensure a fair comparison, CrossGait employs identical backbone networks as feature encoders with these two baseline models. This allows us to isolate the impact of the architectural design and provide a fair comparison. We observed CrossGait underperform GaitBase and LidarGait. Specifically, there was a decrease in Rank-1 accuracy of 5.0\% compared to GaitBase and 1.8\% compared to LidarGait. This decrease in accuracy might be caused by the design of extracting common features across modalities in cross-modality tasks. While this cross-modality alignment helps capture modality-shared gait information, it may lead to a loss of modality-specific details for robust recognition in the single-modality setting.

\noindent \textbf{Cross-modality Evaluation.} Due to the lack of existing cross-modality gait recognition methods in the literature, we compare our proposed CrossGait with five representative methods from the field of visible-infrared cross-modality person re-identification (VI-ReID). Although these five methods have shown promising results in VI-ReID tasks, it is important to note that none of them perform well in cross-modality gait recognition tasks. This highlights the significant gap between cross-modality gait recognition and person re-identification, and it also validates the superiority of our proposed CrossGait. Furthermore, existing gait recognition methods do not provide any results in the cross-modality evaluation protocol, further underscoring the need for novel approaches in the field of cross-modality gait recognition. All methods demonstrate a significant decrease in performance under nighttime conditions, primarily due to the substantial modality variance between low-quality silhouettes and point clouds. The performance decline is primarily because large modality variance between low-quality silhouettes and point clouds. To improve cross-modality gait recognition and mitigate this issue, it is recommended to explore the potential of utilizing raw RGB images from both LiDAR sensors and RGB cameras, instead of relying solely on segmented silhouettes.

\begin{figure*}[t]
  \centering

   \includegraphics[width=0.85\linewidth]{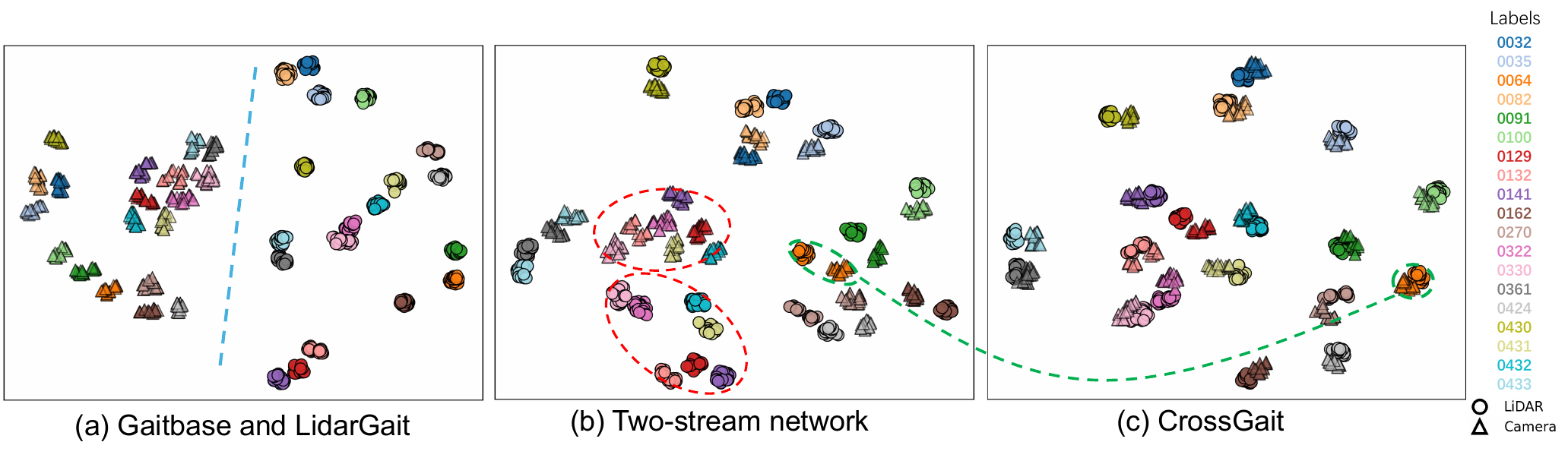}
\vspace{-1em}
   \caption{Feature visualization by t-SNE~\cite{van2008visualizing} of (a) GaitBase~\cite{fan2023opengait} and LidarGait~\cite{shen2022gait}, (b) two-stream network~\cite{wu2017rgb}, and (c) our proposed CrossGait. 
   }

   \label{fig:T-SNE}
   \vspace{-2ex}
\end{figure*}

\begin{figure*}[h]
  \centering

   \includegraphics[width=0.8\linewidth]{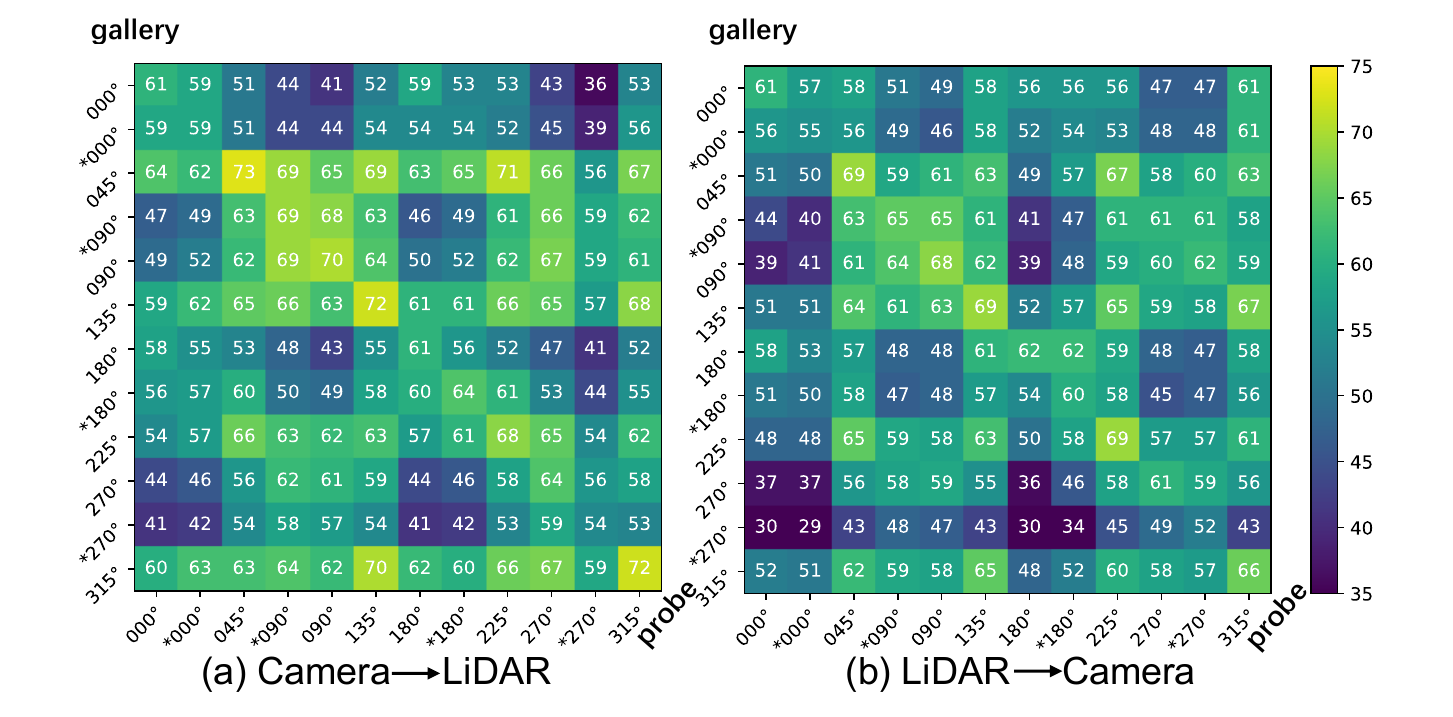}
\vspace{-1em}
   \caption{Cross-view results of our proposed CrossGait on SUSTech1K. Th rank-1 accuracy (\%) is reported. Asterisk (*) denotes viewpoints from longer distances.}
\vspace{-2em}
   \label{fig:crossview}

\end{figure*}



\noindent \textbf{Qualitative Result.} Fig.~\ref{fig:T-SNE} visualizes the feature distribution of the compared methods and the proposed CrossGait. It shows that the features learned by LidarGait and GaitBase are entirely isolated, lacking cross-modality retrieval capability. The two-stream network structure adopts intermediate-level feature alignment to bridge the two modalities, demonstrating its effectiveness but still yielding unsatisfactory results in cross-modality alignment. With a prototypical modality-shared attention module and cross-modality feature adapter, our proposed CrossGait learns a more effective modality-shared feature space, where samples from the same identity/modalities are grouped closely together.

\noindent\textbf{Cross-view Gait Recognition.} We conduct a detailed comparison of cross-view gait recognition in Fig.~\ref{fig:crossview}.
From this comparison, several key observations emerge: (1) Cross-modality recognition encounters challenges at query views of 0°, 90°, and 180°, mirroring the limitations observed in camera-based gait recognition. This suggests that certain angles pose inherent difficulties for capturing reliable gait features. (2) When utilizing LiDAR data as the gallery set, the system demonstrates enhanced cross-perspective capabilities compared to when camera data serves as the gallery. This outcome underscores the importance of 3D attributes in overcoming perspective-related challenges in gait recognition.

\subsection{Ablative Study}

\begin{table}[t]
\centering
    \begin{minipage}[t]{0.44\linewidth}
        \caption{Ablative study on Prototypical Modality-shared Attention Module with different \#$K$ on SUSTech1K.}
        \vspace{-1em}
        \centering
        \resizebox{0.8\textwidth}{!}{%
        \begin{tabular}{c|cccc}
        \hline
                  & \multicolumn{4}{c}{Overall}                                                                      \\ \cline{2-5} 
                  & \multicolumn{2}{c|}{Camera $\rightarrow$ LiDAR} & \multicolumn{2}{c}{LiDAR $\rightarrow$ Camera} \\ \hline
        \#$K$ & \textit{R@1}       & \multicolumn{1}{c|}{\textit{R@5}}      & \textit{R@1}                 & \textit{R@5}            \\ \hline
        0                     & 54.7          & \multicolumn{1}{c|}{78.1}          & 52.1           & 75.9          \\
        1                     & 56.0          & \multicolumn{1}{c|}{79.2}          & 52.9           & 76.3          \\
        2                     & \textbf{56.4} & \multicolumn{1}{c|}{\textbf{79.8}} & \textbf{53.6}  & 77.0 \\
        3                     & 56.3          & \multicolumn{1}{c|}{79.7}          & 53.6           & 77.1          \\
        4                     & 56.3          & \multicolumn{1}{c|}{79.6}          & 53.6          & \textbf{77.2}          \\ \hline
        \end{tabular}%
        \label{Cross-modality Attention Pooling}
        }
    \end{minipage}
    \hfill   
    \begin{minipage}[t]{0.52\linewidth}
        \caption{Ablative study on Cross-modality Feature Adapter (CMFA) on SUSTech1K.}
        \vspace{0.4ex}
        \centering
        \renewcommand{\arraystretch}{1.2}
        \resizebox{1\textwidth}{!}{%
        \begin{tabular}{l|llll}
        \hline
                                        & \multicolumn{4}{c}{Overall}                                             \\ \cline{2-5} 
        Setting & \multicolumn{2}{c|}{Camera $\rightarrow$ LiDAR}    & \multicolumn{2}{c}{LiDAR $\rightarrow$ Camera} \\
                                        & \textit{R@1} & \multicolumn{1}{l|}{\textit{R@5}} & \textit{R@1}          & \textit{R@5 }        \\ \hline
        CrossGait \textit{w/o} CMFA               & 52.8  & \multicolumn{1}{l|}{76.8}  & 50.6           & 74.8          \\
        CrossGait \textit{w} CMFA                      & \textbf{56.4(+3.6)}  & \multicolumn{1}{l|}{\textbf{79.8(+3.0)}}  & \textbf{53.6(+3.0)}           & \textbf{77.0(+2.2)}          \\ \hline
        \end{tabular}%
        \label{Tab:mappinglayer}
        }
    \end{minipage}
\vspace{-1em}
\end{table}


\begin{table}[t]
    \begin{minipage}[t]{0.44\linewidth}
        \caption{Performance of CrossGait utilizing different feature encoders for LiDAR and camera modality.}
        \vspace{-1em}
        \centering
        \renewcommand{\arraystretch}{1.2}
        \resizebox{1\textwidth}{!}{%
        
        \begin{tabular}{cc|cc|cc}
        \hline
        \multicolumn{2}{c|}{Encoder} & \multicolumn{2}{c|}{Camera $\rightarrow$ LiDAR} & \multicolumn{2}{c}{LiDAR $\rightarrow$ Camera} \\ \hline
        LiDAR         & Camera       & \textit{R@1}                         & \textit{R@5}               & \textit{R@1}                    & \textit{R@5}                   \\ \hline
        ResNet-9      & ResNet-18    & 40.6                        & 67.5              & 42.1                   & 69.1                  \\
        ResNet-18     & ResNet-9     & 47.1                        & 74.0              & 46.3                   & 72.1                  \\
        PointNet++    & ResNet-9     & 25.0                        & 52.9              & 25.5                   & 53.2                  \\
        ResNet-9      & ResNet-9     & \textbf{56.4}               & 79.7              & \textbf{53.6}          & \textbf{77.0}         \\ \hline
        \end{tabular}
        \vspace{-4em}
        \label{Tab:DifEncoders}
        }
    \end{minipage}
    \hfill 
    \begin{minipage}[t]{0.52\linewidth}
        \caption{Performance of CrossGait utilizing encoder with different depths on SUSTech1K.}
        \vspace{0.4ex}
        \centering
        \resizebox{0.87\textwidth}{!}{%
        \begin{tabular}{l|llll}
        \hline
        
             & \multicolumn{2}{c|}{Camera $\rightarrow$ LiDAR}                    & \multicolumn{2}{c}{LiDAR $\rightarrow$ Camera} \\ \cline{2-5} 
        Encoder                      & \textit{R@1}         & \multicolumn{1}{c|}{\textit{R@5}}         & \textit{R@1}          & \textit{R@5}         \\ \hline
        ResNet-9               & \textbf{56.4} & \multicolumn{1}{l|}{\textbf{79.7}} & \textbf{53.6}  & \textbf{77.0} \\
        ResNet-18              & 48.8(-7.6)   & \multicolumn{1}{c|}{74.8(-4.9)}   & 45.1(-8.5)    & 71.6(-5.4)   \\
        ResNet-34              & 42.2(-14.2)   & \multicolumn{1}{c|}{70.1(-9.6)}   & 40.0(-13.6)    & 67.4(-9.56)   \\
        ResNet-50              & 41.1(-15.3)  & \multicolumn{1}{c|}{69.1(10.6)}   & 38.6(-15.0)   & 66.6(-10.4)  \\ \hline
        \end{tabular}%
        \label{Tab:backboneselection}
        }
    \end{minipage}
\vspace{-1em}
\end{table}





\noindent\textbf{The Effectiveness of Prototypical Modality-shared Attention Module (PMAM).} We assess the effectiveness of our proposed PMAM in Tab.~\cref{Cross-modality Attention Pooling}. The results of CrossGait without PMAM indicate that the inclusion of PMAM led to a significant improvement in Rank-1 accuracy by 1.8\% for Camera $\rightarrow$ LiDAR matching and 1.5\% for LiDAR $\rightarrow$ Camera matching, compared to the baseline. These results demonstrate that the PMAM successfully bridges the gap between learned features from the two modalities, thereby enhancing the performance of cross-modality recognition.



\noindent\textbf{The Effectiveness of Cross-modality Feature Adapter (CMFA).} We also assess the impact of the CMFA by removing it and applying our contrastive alignment loss directly to align features from the two modalities. The results from Tab.~\ref{Tab:mappinglayer} indicate that the inclusion of the CMFA significantly improves the accuracy of cross-modality recognition. The CMFA enhances the rank-1 accuracy for \textit{Camera $\rightarrow$ LiDAR} by 3.6\% and for \textit{LiDAR $\rightarrow$ Camera} by 3.0\%, validating the effectiveness of the proposed CMFA.


\noindent\textbf{The Selection and Depth of Feature Encoder.} It is also essential to investigate the impact of feature encoders utilized as LiDAR encoders and camera encoders. Tab.~\ref{Tab:DifEncoders}  and Tab.~\ref{Tab:backboneselection} employ different networks as feature encoders, and we observe that: (1) CrossGait achieves better performance when LiDAR features are encoded by a convolutional encoder rather than a point-wise encoder such as PointNet++~\cite{qi2017pointnet++}. We hypothesize that PointNet++ encodes features from raw point clouds, which may result in the extraction of different semantic and structural information. In other words, using a similar convolutional encoder for both the LiDAR and camera branches enables the learning of similar semantic gait patterns, which leads to better performance. (2) Utilizing deeper convolutional layers leads to worse performance. This is due to the limited scale of the training set. With only 250 identities in the SUSTech1K training set, using deeper convolutional layers may result in overfitting.


\noindent\textbf{Generalization to Other Input.} In addition to cross-modality recognition between LiDAR point clouds and camera silhouettes, our proposed CrossGait can also handle other camera-based gait representations, such as praising images and skeleton maps. As shown in Tab.~\ref{Tab:ModalityComparisons}, we observe that the choice of camera-based gait representations has a significant impact on cross-modality recognition performance. Gait representations with shape information, such as silhouettes and parsing images, outperform representations with only structural information, like skeleton maps, by a large margin. This highlights the importance of the differences between representations in cross-modality gait recognition.
To demonstrate the generalization ability of CrossGait to other modalities, we conducted cross-input experiments on diverse gait representations from the camera, including silhouettes, skeleton maps, and parsing images. As illustrated in Tab.~\ref{Tab:cms} and Tab.~\ref{Tab:cmc}, CrossGait exhibits strong generalization ability across these different gait representations. This finding suggests that our approach has the potential to handle future modalities, such as event images, infrared images, or WiFi signals.

\begin{table}[t]

\caption{Cross-retrieval results of CrossGait when utilizing different gait representations on SUSTech1K~\cite{shen2023lidargait}.}
\vspace{-1em}
\centering
\renewcommand{\arraystretch}{1.2}
\resizebox{\textwidth}{!}{%
\begin{tabular}{cc|cccccc|cccccc}
\hline
\multicolumn{2}{c|}{\multirow{2}{*}{Input Modality}} & \multicolumn{6}{c|}{Camera $\rightarrow$ LiDAR}                                                            & \multicolumn{6}{c}{LiDAR $\rightarrow$ Camera}                                                             \\ \cline{3-14} 
\multicolumn{2}{c|}{}                                & \multicolumn{4}{c|}{ProbSequence (R@1 acc)}                                & \multicolumn{2}{c|}{Overall}  & \multicolumn{4}{c|}{ProbSequence (R@1 acc)}                                & \multicolumn{2}{c}{Overall}   \\ \cline{1-2}
\multicolumn{1}{c|}{LiDAR}       & Camera            & Normal        & Clothing      & Umbrella      & \multicolumn{1}{c|}{Night} & \textit{R@1}  & \textit{R@5}  & Normal        & Clothing      & Umbrella      & \multicolumn{1}{c|}{Night} & \textit{R@1}  & \textit{R@5}  \\ \hline
\multicolumn{1}{c|}{Points}      & Silhouettes       & 31.0          & 10.8          & 27.8          & \multicolumn{1}{c|}{3.5}   & 25.0          & 52.9          & 34.9          & 16.0          & 24.8          & \multicolumn{1}{c|}{6.0}   & 25.5          & 53.2          \\
\multicolumn{1}{c|}{Depths}      & Parsings          & 51.8          & 22.4          & 43.8          & \multicolumn{1}{c|}{8.4}   & 45.4          & 71.9          & 53.0          & 24.1          & 30.2          & \multicolumn{1}{c|}{6.8}   & 44.3          & 70.9          \\
\multicolumn{1}{c|}{Depths}      & Skeleton Maps     & 30.3          & 10.4          & 23.1          & \multicolumn{1}{c|}{8.1}   & 24.3          & 50.3          & 27.9          & 13.1          & 11.3          & \multicolumn{1}{c|}{6.7}   & 20.9          & 45.2          \\
\multicolumn{1}{c|}{Depths}      & Silhouettes       & \textbf{62.2} & \textbf{35.4} & \textbf{57.8} & \multicolumn{1}{c|}{10.3}  & \textbf{56.4} & \textbf{79.7} & \textbf{63.2} & \textbf{30.6} & \textbf{38.5} & \multicolumn{1}{c|}{11.8}  & \textbf{53.6} & \textbf{77.0} \\ \hline
\end{tabular}%
\label{Tab:ModalityComparisons}
\vspace{-12ex}
}
\end{table}

\begin{table}[t]
    \centering
    \begin{minipage}[t]{0.48\linewidth}
    \caption{Cross-retrieval of different camera-based gait representations on SUSTech1K ~\cite{shen2023lidargait}.}
        \centering
        \resizebox{1\linewidth}{!}{%
        \begin{tabular}{cc|cc|cc}
        \hline
        \multicolumn{2}{c|}{Input}                      & \multicolumn{2}{c|}{B $\rightarrow$ A} & \multicolumn{2}{c}{A $\rightarrow$ B} \\ \hline
        \multicolumn{1}{c|}{A}          & B             & \textit{R@1}       & \textit{R@5}      & \textit{R@1}      & \textit{R@5}      \\ \hline
        \multicolumn{1}{c|}{Silhouettes} & Parsings       & \textbf{56.4}      & \textbf{80.5}     & \textbf{56.3}     & \textbf{80.6}     \\
        \multicolumn{1}{c|}{Silhouettes} & Skeleton Maps & 22.7               & 47.7              & 21.8              & 47.1              \\
        \multicolumn{1}{c|}{Parsings}    & Skeleton Maps & 24.0               & 50.4              & 22.9              & 48.3              \\ \hline
        \end{tabular}
        \label{Tab:cms}
        }

    \end{minipage}
    \hfill
    \begin{minipage}[t]{0.48\linewidth}
    \caption{Cross-retrieval of different camera-based gait representations on CCPG~\cite{li2023depth}.}
        \centering
        \resizebox{0.87\linewidth}{!}{%
        \begin{tabular}{cc|c|c}
        \hline
        \multicolumn{2}{c|}{Input}                       & B $\rightarrow$ A & A $\rightarrow$ B \\ \hline
        \multicolumn{1}{c|}{A}           & B             & \textit{R@1}                  & \textit{R@1}                  \\ \hline
        \multicolumn{1}{c|}{Silhouettes} & Parsings      & \textbf{64.8}                 & \textbf{65.1}                 \\
        \multicolumn{1}{c|}{Silhouettes} & Skeleton Maps & 29.2                          & 31.3                          \\
        \multicolumn{1}{c|}{Parsings}    & Skeleton Maps & 32.7                          & 33.8                          \\ \hline
        \end{tabular}
        \label{Tab:cmc}
        }
    \end{minipage}
    \vspace{-2em}
 
\end{table}

\vspace{-1em}
\section{Conclusion}
In conclusion, this paper proposes CrossGait, the first cross-modality gait recognition framework that bridges the gap between LiDAR point clouds and camera silhouettes. We introduce the Prototypical Modality-Shared Attention Module (PMAM) to extract common features among different modalities and a cross-modality feature adapter to synchronize features into a cohesive, modality-shared space.
The contributions of this work are threefold. First, we explore cross-modality gait recognition involving LiDAR and a visual camera, which is a novel and unexplored area. Second, we propose CrossGait, a simple yet effective framework that achieves satisfactory performance in retrieving pedestrians across various modalities and even in single-modality settings. Lastly, CrossGait demonstrates its versatility by handling data from different sensors as well as data from the same device but in various formats like human parsing, silhouettes, and skeleton maps.
Overall, this work opens up new possibilities for cross-modality gait recognition and showcases the potential of multi-modal approaches in improving recognition performance across diverse data modalities. Future research can further explore the application of CrossGait to other modalities, such as event images, infrared images, or WiFi signals, and extend its capabilities to handle more complex and challenging recognition tasks.


{\small
\bibliographystyle{ieee_fullname}
\bibliography{egbib}
}

\end{document}